%% file: main.tex
\documentclass[runningheads]{llncs}
\usepackage[T1]{fontenc}

\usepackage{times}
\usepackage{helvet}
\usepackage{courier}
\usepackage{times}
\usepackage{latexsym}
\usepackage{algorithmic}  
\usepackage{amsmath}  
\usepackage{algorithm} 
\usepackage[algo2e]{algorithm2e} 
\usepackage[utf8]{inputenc} 
\usepackage[T1]{fontenc}    
\usepackage{hyperref}       
\usepackage{url}            
\usepackage{booktabs}       
\usepackage{amsfonts}       
\usepackage{nicefrac}       
\usepackage{microtype}      
\usepackage{xcolor}         
\usepackage{makecell}
\usepackage{xcolor,colortbl}
\usepackage{inconsolata}

\usepackage{times}
\usepackage{soul}
\usepackage{url}
\usepackage{amsmath}
\usepackage{booktabs}
\usepackage{algorithm}
\usepackage{algorithmic}
\usepackage{caption}
\usepackage{makecell} 
\usepackage{graphicx}

\usepackage{xcolor,colortbl}

\newcommand{\hide}[1]{}
\newcommand{\attrib}{\textsc{PromptAttrib}}

\usepackage[T1]{fontenc}

\usepackage[utf8]{inputenc}

\usepackage{microtype}

\usepackage{inconsolata}

\begin{document}
%
\title{Prompt-tuning with Attribute Guidance for Low-resource Entity Matching}

\titlerunning{Prompt-tuning with Attribute Guidance for Low-resource Entity Matching}

\author{Lihui Liu\inst{1}\thanks{Lihui Liu is the first author and the corresponding author.\\Code can be found from \url{https://github.com/lihuiliullh/PROMPTATTRIB/tree/main}} \and
Carl Yang\inst{2}}

\authorrunning{Lihui Liu.}

\institute{
Wayne State University, Detroit, MI, USA \and
Emory University, Atlanta, GA, USA
}

\maketitle
\begin{abstract}
Entity Matching (EM) is a significant task involving the determination of the logical relationship between two entities, such as \texttt{Same}, \texttt{Different}, and \texttt{Undecidable}. Traditional approaches to entity matching (EM) heavily depend on supervised learning, which necessitates a vast collection of high-quality labeled data. This labeling process is both time-consuming and costly, limiting the practical application of these methods. Consequently, there is a pressing demand for low-resource EM solutions that can perform effectively with minimal labeled data.
Recently, prompt tuning-based approaches have shown promising results on low-resource entity matching, but they tend to focus solely on entity-level matching, overlooking crucial attribute-level information. Moreover, they lack interpretability and explainability.
To address this limitation, this paper introduces \attrib, a comprehensive solution that tackles entity matching challenges through attribute level prompt tuning and logical reasoning. 
\attrib\ leverages both entity-level and attribute-level prompts to enhance matching accuracy by incorporating valuable contextual information, and it induces the matching label by fuzzy logic formulas. 
By considering attributes, the model gains a deeper understanding of the entities, leading to improved matching results. 
Moreover, \attrib\ incorporates dropout-based contrastive learning on soft prompts, inspired by the SimCSE technique. This further improves the performance of entity matching. Extensive experiments on real-world datasets demonstrate the efficacy of \attrib.
\end{abstract}

\section{Introduction}

\input{001_introduction.tex}

\section{Problem Definition}\label{attrib:problem-definition}

\input{002_problem_definition.tex}

\section{Method}\label{attrib:method}

\input{003_method.tex}

\section{Experiment}\label{attrib:experiment}

\input{004_experiment}

\section{Related Work}\label{attrib:related_work}

\input{005_related_work.tex}

\section{Conclusion}\label{attrib:conclusion}

\input{006_conclusion.tex}

\bibliography{custom,liu}
\bibliographystyle{splncs04}


\end{document}

%% file: 001_introduction.tex
Entity Matching (EM) is a fundamental problem in data management that focuses on determining whether two entity records refer to the same real-world entity. This task carries significant importance in various domains such as question answering ~\cite{tapaswi2016movieqa,bart}, dialog system ~\cite{llama2,alexa,google_lamda}, recommender system ~\cite{wang_recommender} and so on. 
In practical terms, consider a situation where two distinct knowledge databases store customer information from different sources. Due to differences in schema and data formats, effectively combining this information becomes a challenging task. However, by leveraging entity matching techniques, it becomes possible to identify matching customer records across knowledge databases, facilitating accurate data merging and linking. Thus, the application of entity matching plays a crucial role in improving data quality and enabling seamless information consolidation across diverse sources.

Conventional entity matching (EM) techniques rely extensively on abundant high-quality labeled data, which is often scarce or unavailable. Therefore, there is a growing need for low-resource EM methods that can achieve robust performance with minimal labeled data.
Prompt tuning is a promising way for training and fine-tuning language models to tackle low-resource entity matching. By carefully crafting and adjusting the initial prompt given to the model, researchers and developers can guide the model's behavior and steer it towards desired outcomes. 
Existing prompt tuning based methods for low-resource entity matching, such as promptEM ~\cite{promptEM}, commonly serialize each entity into a sentence, and a prompt template is used to generate the input for the language model. Despite these approaches have shown good performance in many scenarios, such prompt tuning based models function like a black-box and lack interpretability. Moreover, they consider only the entity-level information and overlook the crucial attribute information associated with each entity.

However, attribute information plays a significant role in accurately determining entity matches because it contains more fine-grained details. It can help us explain why two entities are the \texttt{Same}, \texttt{Different}, or \texttt{Ambiguous}. For example, Michael Jordan with the occupation "basketball player" is different from Michael Jordan with the occupation "computer scientist". 
Exploring innovative ways to incorporate attribute-level information holds promise for enhancing the robustness and reliability of entity matching systems.

In addition, existing prompt tuning methods typically assume that the input prompt embedding is well-trained during the training process, without explicitly making the representations of different inputs distinguishable from each other. However, high-quality entity/attribute representations are crucial for the success of entity matching. If two entities/attributes are different, their representations should be easily distinguishable. 
Taking inspiration from SimCSE, a technique that utilizes dropout-based contrastive learning on language models to enhance representation learning quality, we propose applying dropout-based contrastive learning on prompt tuning. This approach aims to increase the model's resilience to variations in input prompts by introducing dropout mechanisms during the training phase. By incorporating dropout-based contrastive learning on soft prompts, we seek to improve the overall performance of the entity matching model, ultimately leading to more accurate and consistent results.

In this paper, we present \attrib, a comprehensive solution consisting of two main components. First, \attrib\ leverages both entity-level prompts and attribute-level prompts to effectively address the challenges related to low-resource entity matching. By incorporating fuzzy logic reasoning, the model predicts entity-level matches based on the logical relationships between different attribute information, making the prediction results more explainable.
Second, \attrib\ employs dropout-based contrastive learning on soft prompts. This technique enhances the model's performance by encouraging distinguishable feature representations and reducing overfitting. Through dropout-based contrastive learning, the model learns to generalize better and captures more nuanced relationships between entities and attributes.

In summary, the main contributions of this paper are:
\begin{itemize}
    \item \textbf{Algorithm}: We propose \attrib, which utilizes entity-level prompts, attribute-level prompts, fuzzy logic reasoning, and soft token contrastive learning to tackle low-resource entity matching.
    \item \textbf{Empirical Evaluations}: We conducted extensive experiments on several real-world datasets. The results of our experiments demonstrate the effectiveness of \attrib.
\end{itemize}


%% file: 002_problem_definition.tex
Low-resource entity matching (EM) involves the identification of pairs of data entries from two collections that refer to the same real-world entity with limited labeled training datapoints. 
Formally, given two data sources $E_A$ and $E_B$, we assign a binary label $y \in \{0, 1\}$ to each candidate pair ($e_a$, $e_b$) $\in$ $E_A \times E_B$. Here, $y = 1$ indicates a true match, while $y = 0$ represents a mismatched pair.

\begin{problem}{Low-resource Entity Matching: }

	\textbf{Given:} (1) an entity $e_a$ from data source $E_A$, (2) an entity $e_b$ from data source $E_B$, (3) limited labeled training datapoints.
	
	\textbf{Output:} a binary label $y \in \{0, 1\}$ for the entity pair ($e_a$, $e_b$)
\end{problem}

\begin{figure}
    \centering
    \includegraphics[width=0.65\textwidth]{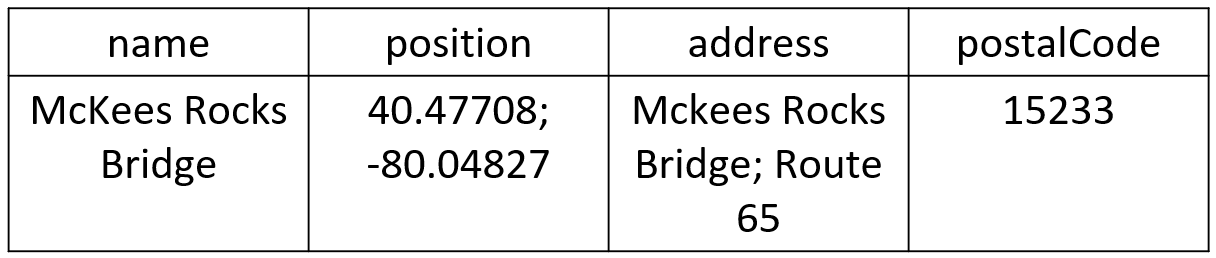}
    \caption{Example.
    }
    \vspace{-1\baselineskip}
    \label{attrib:example}
\end{figure}

\noindent \textbf{Serializing:} The matching problem can be effectively solved by formulating it as a sequence classification task. First, entity pairs are serialized to sequences, and then, a pre-trained LM is finetuned to solve the task. 
An entity with $n$ attributes can be denoted as $e = \{\text{attr}_i, \text{val}_i\}_{i \in [1,n]}$, where $\text{attr}_i$ is the attribute name and $\text{val}_i$ is the corresponding attribute value. Then the serialization ~\cite{promptEM} is denoted as:
\begin{align*}
\text{serialize}(e) ::= [\text{COL}] \text{attr}_1 [\text{VAL}] \text{val}_1 \ldots 
[\text{COL}] \text{attr}_n [\text{VAL}] \text{val}_n
\end{align*}
where $[\text{COL}]$ and $[\text{VAL}]$ are two special tags indicating the start of attribute names and values, respectively. Taking the relational entity in Figure ~\ref{attrib:example} as an example, we serialize it as:
\begin{align*}
[\text{COL}] \text{name} [\text{VAL}] \text{McKees Rocks Bridge} \ldots 
[\text{COL}] \text{postalCode} [\text{VAL}] \text{15233} 
\end{align*}

\noindent \textbf{Prompt-based Tuning:}
Prompt-based tuning has been proposed to apply cloze-style tasks to tune LMs. Formally, we define a label word set \(V_{\mathbf{y}} = \{w_1, \ldots, w_m\}\). \(V_{\mathbf{y}}\) is a subset of the vocabulary \(V\) of the LM, i.e., \(V_{\mathbf{y}} \subseteq V\). We get an overall dictionary \(V^*\) by taking the union of the dictionary corresponding to each label. Another primary component of prompt-based tuning is a prompt template \(T(\cdot)\), which modifies the original input \(x\) into a prompt input \(T(x)\) by adding a set of additional tokens in \(x\). Generally, a token \([MASK]\) is added for LMs to predict the missing label word \(w \in V_{\mathbf{y}}\) using \(T(x)\). Thus, in prompt-tuning, a classification problem is transferred into a masked language modeling problem: $p(y \in Y | x) = p([MASK] = w \in V_{\mathbf{y}} | T(x))$, 
where $Y$ is the label set.

%% file: 003_method.tex
\begin{figure}[t!]
    \centering
    \includegraphics[width=0.85\textwidth]{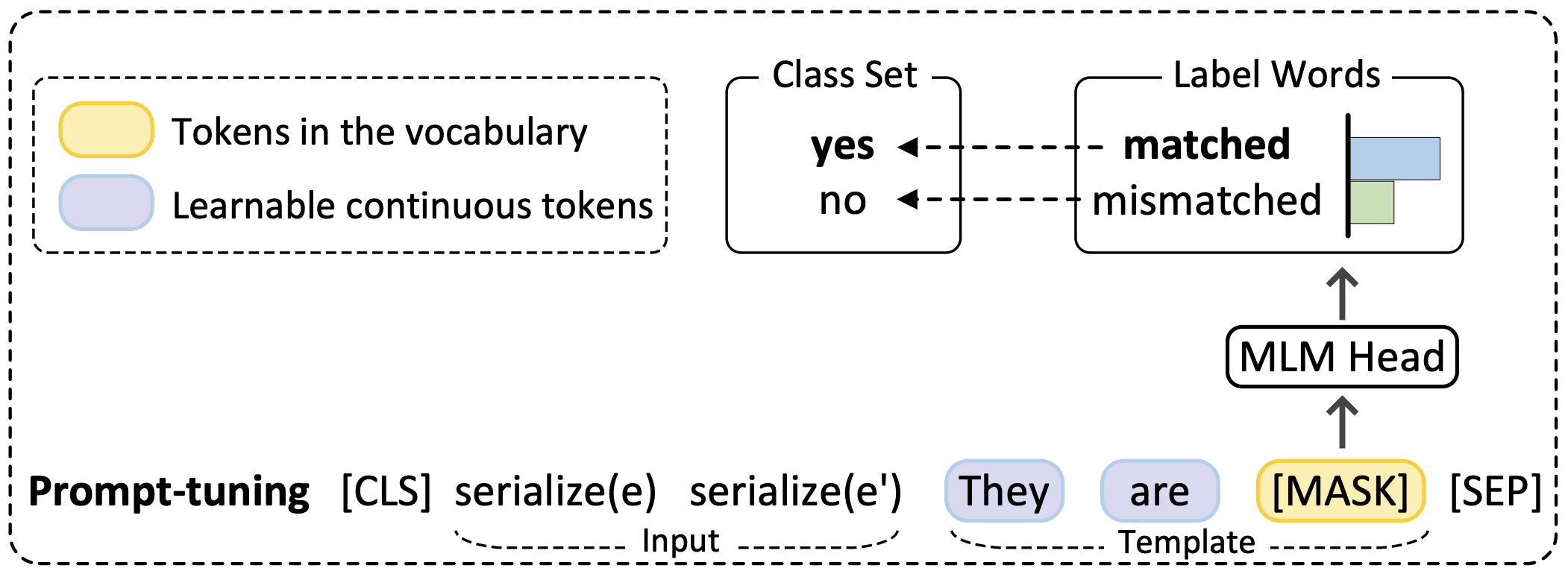}
    \caption{
    The illustration of prompt-tuning. The blue rectangles in the figure are special prompt tokens, whose parameters are initialized and learnable during prompt-tuning. 
    }
    \label{attrib:promptEM}
\end{figure}

In this section, we detail how to utilize prompt-tuning to deal with EM. We first design EM-specific prompt templates and label words, and then, we describe the training and inference process. 

\subsection{Prompt Template}\label{attrib:prompt_template}

To cast the EM problem as a prompt-tuning one, we first design suitable prompt templates (i.e., hard-encoding templates and continuous templates) and label words set (to consider general binary relationship) following ~\cite{promptEM}. The hard-encoding templates and continuous templates are the same as those in ~\cite{promptEM}, except that we use a different prompt template and label words set. 
Inspired by ~\cite{promptEM}, given each candidate pair \(x = (e, e')\), we construct the following templates\footnote{Note that various prompts can be used in the experiments, we provide only two simple examples here.}:
\begin{align*}
T_1(x) &= \text{{Are}} \ \text{{serialize}}(e) \ \text{{and}} \ \text{{serialize}}(e')\ \text{{the}} \ [MASK] \\
T_2(x) &= \text{{serialize}}(e) \ \text{{is}} \ [MASK] \ \text{{to}} \ \text{{serialize}}(e')
\end{align*}

Since the goal of prompt construction is to optimize task performance rather than readability, prompts don't need to be human-interpretable. Continuous prompts, embedded directly in the model's space, address this need. We use P-tuning ~\cite{promptEM}, where trainable prompt tokens are integrated into the embedding layer and processed by BiLSTM to capture their interactions. This approach allows the model to learn more effective prompts beyond its original vocabulary. An example is shown in Figure ~\ref{attrib:promptEM}.

\begin{figure*}
    \centering
    \includegraphics[width=0.99\textwidth]{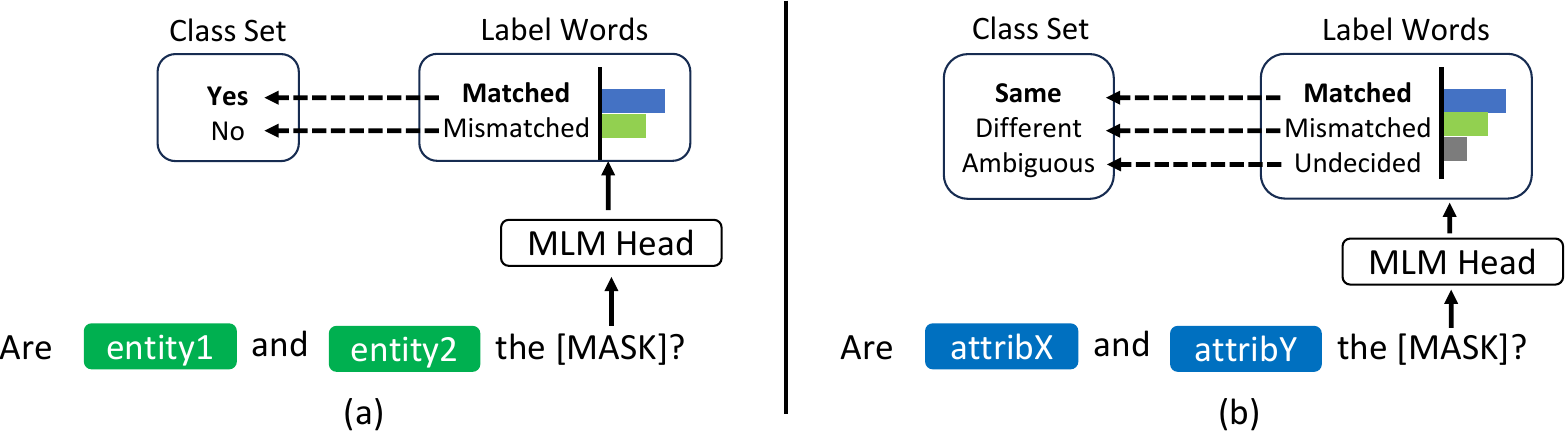}
    \caption{The entity level prompt tuning and the attribute level prompt tuning. 
    }
    \label{entity_attrib_prompt}
    \vspace{-1.5\baselineskip}
\end{figure*}

\subsubsection{Label Words Set.}
In addition to designing templates, another primary component is to design the set of label words. Given two entities \(e_1\) and \(e_2\), it is often more informative to classify the relationship between them rather than simply determining if they are identical. Specifically, we aim to determine if the entities are relevant to each other, which encompasses a broader range of relationships beyond mere matching.
For a general binary relationship, we map the label \(y = \text{yes}\) into a set 
$V_y = \{\text{matched, similar, relevant}\}$.
Similarly, the label \(y = \text{no}\) is mapped to a set 
$V_y = \{\text{mismatched, different, irrelevant}\}$.
This approach allows for a more nuanced classification that captures various degrees of relevance and irrelevance between the entities.

\subsection{Method Details}

\subsubsection{Entity Level Prompt-tuning}
Inspired by existing prompt tuning-based methods, such as PromptEM~\cite{promptEM,ditto}, we consider checking whether two entities are the same by prompt-tuning. More specifically, given two entities, we first serialize each entity into text using the method introduced in Subsection~\ref{attrib:prompt_template}. Subsequently, these two serialized texts are used as input for the prompt template, as depicted in Figure~\ref{entity_attrib_prompt}(a). The resulting context is then fed as input to the large language model. The output embedding of the special token \textbf{[MASK]} is utilized by the \textbf{MLM Head} to predict the results.
The label words set used in entity level prompt tuning is the same as the label words set introduced in Subsection~\ref{attrib:prompt_template}. This means we map the label \(y = \text{yes}\) to a set 
$V_y = \{\text{matched, similar, relevant}\}$.
Similarly, the label \(y = \text{no}\) is mapped to a set 
$V_y = \{\text{mismatched, different, irrelevant}\}$.

\subsection{Attribute Level Prompt-tuning}
Existing prompt tuning-based methods primarily focus on considering only the entity-level information of two entities. However, attribute information is often of significant importance, as it can provide essential context to the system. To leverage this attribute information effectively, one potential approach is to integrate it into the prompt. In contrast to entity-level prompts that include only one \textbf{[MASK]} token, the attribute prompt contains both entity-level and attribute-level information, featuring multiple \textbf{[MASK]} tokens in the template. The first \textbf{[MASK]} is utilized to predict the entity-level information, while the subsequent \textbf{[MASK]} tokens are employed to predict the attribute-level information. 

Despite the fact that the above idea is intuitive, certain entities may contain an extensive amount of information. For instance, the \textbf{Description} field of a product, such as a "computer," can be very lengthy, even exceeding 512 words. This situation poses a challenge as the combined length of the enhanced prompt with attribute information would exceed the model's maximum input length, necessitating truncation of the text. To address this issue, we separate the entity-level prompt from the attribute-level prompt, as illustrated in Figure~\ref{entity_attrib_prompt}(b). 
Unlike entity-level prompt tuning, the label words set used in attribute-level prompt tuning contains three categories: \texttt{Same}, \texttt{Different}, and \texttt{Ambiguous}. More specifically, we map the label \(y = \text{Same}\) to a set 
$V_y = \{\text{same, similar, positive}\}$,
the label \(y = \text{Different}\) is mapped to a set 
$V_y = \{\text{mismatched, different, irrelevant}\}$,
and the label \(y = \text{Ambiguous}\) is mapped to a set 
$V_y = \{\text{uncertain, unclear, neutral}\}$.

\subsection{Fuzzy Logic Reasoning}

Because each entity comprises multiple attributes, entity-level matching can be logically induced from attribute-level matching. Based on the definition of entity matching, we have developed the following induction rules.

\noindent \textbf{Same Rule:} It is evident that if two entities are the same, then all their attributes should also be the same, implying that all paired attributes should be labeled as \texttt{Same}.
Let $\{(a_k^1, a_k^2)\}_{k=1}^K$ be all attribute pairs. Then, we induce a attribute-level \texttt{Same} score by
\begin{align*}
S(\texttt{Same}|e_1, e_2) = \left[ \prod_{k=1}^{K} P(\texttt{Same}|a_k^1, a_k^2) \right]^{\frac{1}{K}}
\end{align*}
This works in a fuzzy logic fashion ~\cite{fuzzy_logic}, deciding whether the entity-level label should be \texttt{Same} by considering the average of attribute level predictions. Here, we use the geometric mean because it is biased towards low scores. In other words, if there exists one attribute pair with a low \texttt{Same} score, then the chance of the entity label being \texttt{Same} is also low. 

\noindent \textbf{Difference Rule:} Two entities are \texttt{Different} if there exists (at least) one paired attribute labeled as \texttt{Different}. The fuzzy logic version of this induction rule is given by

{\small
\begin{align*}
S(\texttt{Different}|e_1, e_2) = \max_{k=1,\ldots,K} \ P(\texttt{Different}|a_k^1, a_k^2)
\end{align*}
}

Here, the \text{max} operator is used in the induction because the \texttt{Different} rule is an existential statement, i.e., there exist(s) $\cdots$.

\noindent \textbf{Ambiguous Rule:} Two entities are \texttt{Ambiguous} if there exists (at least) one \texttt{Ambiguous}  attribute pair, but there does not exist any \texttt{Different} attribute pair. The fuzzy logic formula is

{\small
\begin{align*}
&S(\texttt{Ambiguous} \mid e_1, e_2) = \left[ \max_{k=1 \ldots K'} P(\texttt{Ambiguous} \mid a_k^1, a_k^2) \right] \left[1 - S(\texttt{Different} \mid e_1, e_2) \right]
\end{align*}
}

The first factor determines whether there exists a \texttt{Ambiguous} attribute pair. The second factor evaluates the negation of "at least one different attribute" as suggested in the second clause of the Rule for \texttt{Ambiguous}.

Finally, we normalize the scores into probabilities by dividing the sum, since all the scores are already positive. 
This is given by $P(L | \cdot) = Z^{-1} S(L | \cdot)$,
where \( L \in \{ \texttt{Same}, \texttt{Different}, \texttt{Ambiguous} \} \), and \( Z = S(\texttt{Same} | \cdot) + S(\texttt{Different} | \cdot) + S(\texttt{Ambiguous} | \cdot) \) is the normalizing factor.

During training process, we use cross-entropy loss to train our  model by minimizing
$- \log P(t | \cdot)$ where \( t \in \{ \texttt{Same}, \texttt{Different}, \texttt{Ambiguous} \} \) is the ground truth entity-level label. During the experiment, we categorize \texttt{Ambiguous} based on various datasets as either \texttt{Same} or \texttt{Different}.

\subsection{Soft Token Contrastive Learning}

Current methods for tuning prompts typically assume that the initial soft prompt is well-established during training and that the model's learned representation is sufficiently distinguishable. However, this assumption often does not hold, particularly in low-resource scenarios. High-quality representations of entities and attributes significantly influence performance. To tackle this issue, drawing inspiration from techniques such as SimCSE ~\cite{gao2021simcse}, we propose integrating contrastive learning into soft prompts to enhance representation learning.
\vspace{-1.5\baselineskip}
\begin{figure}
    \centering
    \includegraphics[width=0.4\textwidth]{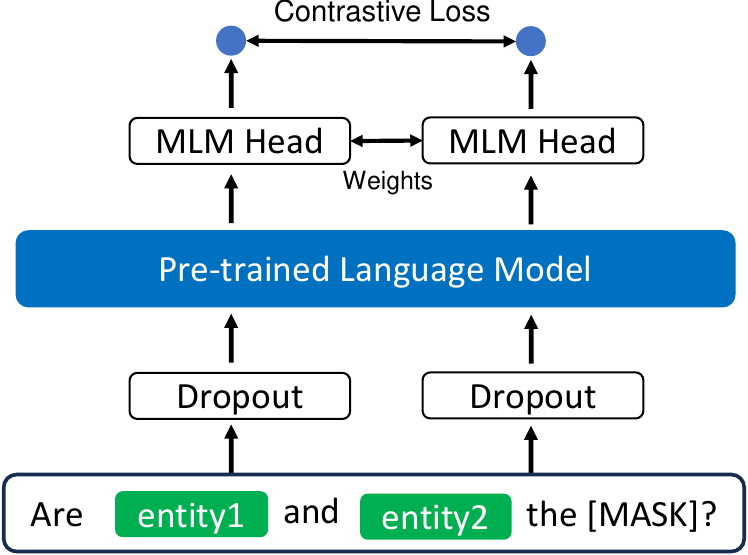}
    \caption{Contrastive learning framework. Dropout is applied to the entire input embedding.
    }
    \label{attrib:contrastive_framework}
    \vspace{-1.5\baselineskip}
\end{figure}

The idea of contrastive learning is to enhance representation learning by training a model to distinguish between similar and dissimilar instances in the data. By maximizing the similarity of representations from similar instances and minimizing that of dissimilar ones, contrastive learning fosters the creation of more meaningful and discriminative feature representations.

In \attrib, given an entity $e_i$, we construct its positive sampling data by randomly applying dropout masks to the soft prompt token inside $\text{{serialize}}(e_i)$, as introduced in the previous sections, resulting in $e_i^+$. This approach has been utilized in various models; for instance, in the standard training of Transformers~\cite{transformer}, dropout masks are applied to fully-connected layers and attention probabilities (typically $p = 0.1$) to enhance model robustness.
After obtaining the positive data point, we pass it through the prompt-tuning model $z = f_{\theta}(e_i^+)$ to obtain its embedding $z$. This process is repeated twice using different dropout masks to generate two positive embeddings $z_1$ and $z_2$, and the training objective becomes: $L = \| z_1 - z_2 \|_2$.

Note that unlike SimCSE, which applies dropout to the language model parameters, we keep the language model unchanged because modifying parameters in language models can be time-consuming. Instead, we apply dropout to the input soft tokens within the prompt. This approach ensures that even if the input prompt context remains constant, applying dropout multiple times yields distinct inputs. This concept bears similarity to siamese networks~\cite{sentencebert}, albeit focusing solely on positive pairs in the contrastive loss. The final framework of the model is shown in Figure ~\ref{attrib:contrastive_framework}











%% file: 004_experiment.tex
In this research, we thoroughly evaluated our method using several datasets that represent diverse and realistic entity-matching scenarios. We used four main datasets. Geo-heter~\cite{promptEM} contains 194,089 geospatial entities collected from platforms such as Yelp, Foursquare, and OpenStreetMap. Cameras~\cite{ditto} and Computers~\cite{ditto} are subsets of the large WDC product dataset, which includes millions of product offers from various e-commerce websites; these subsets focus specifically on camera-related and computer-related entities, including their models, brands, and specifications. Finally, the ISWC dataset~\cite{ditto} includes entities from the International Semantic Web Conference, such as papers, authors, venues, and publication years. Together, these datasets provide a broad and practical testbed for evaluating our approach.
\vspace{-1.5\baselineskip}
\begin{table*}
	\centering
	\caption{The entity matching performance of different methods. F means F1-score, P means average precision, A means Accuracy.}
	\begin{tabular}{|c|c|c|c|c|c|c|c|c|c|c|c|c|c|c|c|c|c|c|c|c|}
	\hline
    Dataset & \multicolumn{3}{c|}{geo-heter}  &  \multicolumn{3}{c|}{cameras}  &  \multicolumn{3}{c|}{computers} & \multicolumn{3}{c|}{ISWC} \\ \hline
	        & F & P  & A & F & P  & A & F & P  & A & F & P  & A \\ \hline
	SentenceBERT & 58.8 & 53.8 &  64.8 &  38.3  &   21.4  &   21.3  &   39.4  &   35.6  &   57.5  &   66.2  &   68.0  &   75.5 \\ \hline
    DeepMatcher & 43.8 & 28.9 & 72.1 & 39.2  &   22.4  &   23.0  &   42.8  &   37.1  &   58.9  &   69.1  &   70.6  &   76.3 \\ \hline
	Ditto   & 3.6  &   45.5 &  71.3 &  42.8  & 27.3 &  27.3  &  44.9  & 40.3 &  62.1  &  72.1  & 72.7 &  80.6 \\ \hline
	PromptEM  & 78.5 &     84.5   & 85.5 &  35.4 & 29.4   & 68.8   &  49.5 & 45.2   & 68.4  & 76.4 & 73.6 & 81.9 \\ \hline
    \rowcolor{lightgray} \attrib  & 81.1 & 85.4   & 88.9  & 45.5 & 45.6   & 72.2  &  47.7 & 49.5   & 72.1 &  77.5 &  79.6   & 82.5  \\ \hline
	\end{tabular}
        \label{attrib:performance}
        \vspace{-1\baselineskip}
\end{table*}

The details of these datasets can be found in~\cite{promptEM} and~\cite{ditto}. In our experiments, we use only 5\% of the labeled training data to simulate a low-resource entity matching setting. To evaluate our approach, we compare it against several strong baseline methods. SentenceBERT~\cite{sentencebert} introduces a siamese architecture built on pretrained language models for sentence matching, which can also be applied to entity matching. DeepMatcher~\cite{deepmatcher} is a traditional entity matching framework that uses recurrent neural networks to aggregate attribute values and align their representations. Ditto~\cite{ditto} is a state-of-the-art approach that fine-tunes a pretrained language model using domain knowledge, TF-IDF summarization, and data augmentation. Finally, PromptEM~\cite{promptEM} adapts prompt-based tuning for entity matching by reformulating the task as a cloze-style masked language modeling problem. Together, these baselines provide a comprehensive comparison for assessing the effectiveness of our method.

All experiments are conducted on a machine with an Intel(R) Xeon(R) Gold 6240R CPU, 1510 GB memory, and an NVIDIA-SMI Tesla V100-SXM2 GPU.

\subsection{Entity Matching Performance}

In this subsection, we test the performance of different baseline methods on entity matching task. We use metrics: F1-score, Average Precision and Accuracy to measure the performances of all the baselines.

Table ~\ref{attrib:performance} shows the results. As we can see, for the geo-heter dataset, \attrib\ outperforms both Ditto and PromptEM, achieving the highest F1-score (81.1\%), Average Precision (85.3\%), and Accuracy (88.9\%). PromptEM also performs well with an F1-score of 78.5\%, while Ditto has the lowest F1-score of 3.6\%.

In the cameras dataset, \attrib\ has the highest F1-score (45.5\%), Average Precision (45.6\%) and Accuracy (72.2\%), while PromptEM achieves the second highest Accuracy of 0.688 compared to other methods.
For the computers dataset, \attrib\ again outperforms the other methods, achieving the highest F1-score (47.7\%), Average Precision (49.5\%) and Accuracy (72.1\%). PromptEM follows closely with an F1-score of 49.5\% and an Average Precision of 45.2\%. Ditto lags behind with an F1-score of 44.9\%.
In the ISWC dataset, \attrib\ also performs the best, obtaining the highest F1-score (77.5\%), Average Precision (79.6\%) and Accuracy (82.5\%). PromptEM closely follows with an F1-score of 76.4\% and, Average Precision of 73.6\% and Accuracy 81.9\%. Ditto achieves an F1-score of 72.1\%.
Overall, the results indicate that \attrib\ consistently outperforms other baselines across the different datasets, showcasing its effectiveness in entity matching tasks. 

\subsection{Performance of Different Language Models}

We also conducted experiments using various large language models as the backbone of the prompt tuning model. Specifically, we evaluated six models, including Bert ~\cite{bert}, Roberta ~\cite{liu2019roberta}, Albert ~\cite{albert}, GPT2 ~\cite{gpt2}, Roberta-large, and Albert-large. The performance results are presented in Table ~\ref{attrib:llm}.

\begin{table}
	\centering
	\caption{The entity matching performance of different language models as backbone. Dataset is geo-heter.}
    \setlength{\tabcolsep}{16pt}
	\begin{tabular}{|c|c|c|c|c|c|c|}
	\hline
	        & F & P  & A \\ \hline
	BERT   & 77.2  & 73.5 &  83.4  \\ \hline
	Roberta  & 79.5 & 78.2   & 86.4 \\ \hline
    Albert  & 78.7 & 80.2   & 86.6 \\ \hline
    GPT2   & 64.7  & 68.4 &  79.7  \\ \hline
    Roberta-large  & \textbf{81.1} & \textbf{85.4} & \textbf{88.9} \\ \hline
    Albert-large  & 79.9 & 83.1   & 86.2 \\ \hline
	\end{tabular}
        \label{attrib:llm}
        \vspace{-1.5\baselineskip}
\end{table}

As we can see, the results show that among the tested language models, Roberta-large achieved the highest F1-score (81.1\%), Average Precision (85.4\%) and Accuracy (88.9\%), closely followed by Albert-large (F1-score: 79.9\%, Average Precision: 83.1\%, and Accuracy: 86.2\%).
Roberta also performs well with an F1-score of 79.5\% and an Average Precision of 78.2\%. BERT obtained an F1-score of 77.2\% and an Average Precision of 73.5\%.
GPT2 achieved a lower F1-score of 64.7\% and an Average Precision of 68.4\% compared to the other language models.
In terms of Accuracy, Roberta-large again stands out with a value of 88.9\%, followed closely by Albert (Accuracy: 86.6\%) and Albert-large (Accuracy: 86.2\%).
Overall, the results demonstrate that language models, such as Roberta-large and Albert-large, tend to perform better than smaller ones like BERT and GPT2, indicating the importance of model size and capacity in improving entity matching performance. 


\subsection{Ablation Study}
In this subsection, we examine the effectiveness of integrating contrastive learning with attribute-level prompt tuning. We evaluate how this combination improves model's overall performance, particularly in low-resource scenarios. By analyzing various metrics and comparing results under different scenarios, we aim to demonstrate their advantages.

\begin{table}
	\centering
	\caption{The entity matching performance with contrastive learning under different dropout ratios. Dataset is cameras.}
	\setlength{\tabcolsep}{12pt}
    \begin{tabular}{|c|c|c|c|c|c|c|}
	\hline
	Model & F & P & A \\ \hline
    Ditto & 42.9 & 27.3 & 27.3 \\ \hline
    PromptEM & 42.3 & 31.7 & 71.3 \\ \hline
    \attrib\ No & 40.4 & 39.7 & 71.7 \\ \hline
    \attrib\ 0.35 & 45.5 & 45.6 & 72.3 \\ \hline
    \attrib\ 0.4 & 37.6 & 39.0 & 70.9 \\ \hline
    \attrib\ 0.45 & 39.1 & 39.5 & 73.1 \\ \hline
	\end{tabular}
        \label{attrib:dropout}
    \vspace{-1.5\baselineskip}
\end{table}

 \textbf{Dropout ratio.} Table~\ref{attrib:dropout} shows the entity matching performance with contrastive learning under different dropout ratios, 
where ``\attrib\ 0.35" means the dropout ratio is 0.35 and ``\attrib\ No" means without dropout. 
As we can see, \attrib\ with a dropout ratio of 0.35 achieves the highest F-score (45.5\%) and average precision (45.6\%), indicating that this ratio provides the best balance between regularization and information retention. Comparing ``\attrib\ No" and \attrib\ with different dropout ratios, it is evident that applying dropout generally improves performance. For instance, ``\attrib\ No" achieves an F-score of 40.4\%, while ``\attrib\ 0.35" achieves a significantly higher F-score of 45.5\%. This improvement suggests that dropout helps in preventing overfitting and enhances the model's ability to generalize from the training data. Interestingly, \attrib\ with a dropout ratio of 0.4 and 0.45 shows a decrease in F-score (37.6\% and 39.1\%, respectively) compared to ``\attrib\ 0.35", indicating that too much dropout can degrade performance. However, the accuracy (A) metric is highest for ``\attrib\ 0.45" at 73.1\%, suggesting that while the F-score and average precision metrics are sensitive to dropout ratio, while accuracy still benefit from higher dropout ratio.

%% file: 005_related_work.tex
\paragraph{Entity Matching:}
Knowledge graph reasoning has been studied for a long time
~\cite{liu2019g,liu2021neural,liu2021kompare}
~\cite{liu2022joint,liu2022comparative,liu2022knowledge,liu2023knowledge,liu2016brps,liu2024logic,liu2024can,liu2024new,liu2024conversational,liu2025transnet,liu2025neural,liu2025few,liu2025monte,liu2025hyperkgr,liu2025mixrag,liu2024knowledge,liu2025unifying,liu2026neural,liuneural,liu2026accurate,liu2026accurate2,liu2026ambiguous,liu2026ambiguous2}.
Entity Matching (EM) is a critical task in knowledge graph reasoning, essential for ensuring data quality and consistency. 
Various methods have been explored, including rule-based approaches, crowdsourcing techniques, and traditional machine learning (ML) models. Recently, deep learning (DL) has shown significant promise in EM. For instance, DeepER ~\cite{deeper} employs neural networks to extract features and treats EM as a classification problem. Similarly, DeepMatcher ~\cite{deepmatcher} provides a comprehensive DL framework for EM. However, DL approaches often require substantial labeled training data, which is costly and impractical.
To address this, Ditto ~\cite{ditto} leverages pre-trained language models and incorporates data augmentation to improve EM performance with less training data. Additionally, Rotom ~\cite{rotom} enhances EM tasks by integrating multiple data augmentation operators, while DADER ~\cite{dader} advances EM through domain adaptation techniques. Other strategies, such as information fusion, active learning, and transfer learning, have also been investigated to boost EM performance.



\paragraph{Prompt Tuning:}
Despite the success of fine-tuning pre-trained language models (LMs) ~\cite{liu2019roberta,bert,albert}, the gap between pre-training and fine-tuning objectives limits the full potential of pre-trained knowledge. The introduction of GPT-3 ~\cite{gpt3} sparked interest in prompt-tuning ~\cite{prompt_tuning0}, which uses hand-crafted prompts to achieve strong performance on various tasks, particularly in low-resource settings. Building on GPT-3, many hand-encoded prompts have been explored.
Recently, methods like continuous prompts ~\cite{prompt_tuning1,prompt_tuning2} have emerged, reducing the need for manual prompt design and improving prompt expressiveness. Prompt-tuning has led to advancements in natural language inference and entity typing.

%% file: 006_conclusion.tex
In this paper, we propose PromptAttrib, a comprehensive solution with two main components for effective entity matching. The first part leverages entity-level prompts and attribute-level prompts to address challenges in entity matching. By incorporating these prompts, the model gains valuable contextual information, enhancing accuracy. In the second part, PromptAttrib employs dropout-based contrastive learning on soft prompts, promoting robust feature representations and reducing overfitting. This approach enables better generalization and captures nuanced entity-attribute relationships. Experimental results demonstrate the efficacy of PromptAttrib, showcasing significant improvements in entity matching performance.

\section{Acknowledge}

This research was partially supported by the the internal funds and GPU servers provided by the Computer Science Department of Emory University, the US National Science Foundation under Award Numbers 2442172, 2312502, 2319449, and the US National Institutes of Health under Award Numbers K25DK135913, RF1NS139325, R01DK143456 and U18DP006922.



